\documentclass[letterpaper, 10 pt, conference]{ieeeconf}  

\IEEEoverridecommandlockouts                              

\usepackage[caption=false]{subfig}

\usepackage{graphicx} 
\usepackage{amsmath} 
\usepackage{color}
\usepackage[makeroom]{cancel}
\usepackage{hyperref}

\renewcommand{\eqref}[1]{Equation~(\ref{#1})}
\newcommand{\figref}[1]{Figure~\ref{#1}}

\renewcommand{\baselinestretch}{0.9}
\makeatletter
\let\NAT@parse\undefined
\makeatother
\usepackage[numbers, sort]{natbib}

\title{\LARGE \bf
Why did the Robot Cross the Road? - 
Learning from Multi-Modal Sensor Data for Autonomous Road Crossing
}

\author{Noha Radwan \and Wera Winterhalter \and Christian Dornhege \and Wolfram Burgard
\thanks{All authors are with the University of Freiburg, Institute of Computer Science, 79110, Germany.
        }%
}

\begin{document}

\onecolumn
{\Large

\textcopyright IEEE. Personal use of this material is permitted. Permission from IEEE must be obtained
for all other uses, in any current or future media, including reprinting/republishing this material
for advertising or promotional purposes, creating new collective works, for resale or redistribution
to servers or lists, or reuse of any copyrighted component of this work in other works.\\

%
Pre-print of article that will appear at the 2017 IEEE International Conference on Intelligent
Robots and Systems.\\

Please cite this paper as:\\
N. Radwan, W. Winterhalter, C. Dornhege and W. Burgard, "Why did the Robot Cross the Road? - Learning from Multi-Modal Sensor Data for Autonomous Road Crossing" in Intelligent Robots and Systems (IROS), 2017 IEEE/RSJ International Conference on. IEEE, 2017.\\

bibtex:\\
@inproceedings$\lbrace$ radwan17iros,\\
author = $\lbrace$ Noha Radwan and Wera Winterhalter and Christian Dornhege and Wolfram Burgard$\rbrace$,\\
booktitle = $\lbrace$ Intelligent Robots and Systems (IROS), 2017 IEEE/RSJ International Conference on$\rbrace$,\\
organization = $\lbrace$IEEE$\rbrace$,\\
title = $\lbrace$ $\lbrace$Why Did the Robot Cross the Road? - Learning from Multi-Modal Sensor Data for Autonomous Road Crossing$\rbrace$ $\rbrace$,\\
year = $\lbrace$2017$\rbrace$ \\
$\rbrace$
}
\twocolumn

\maketitle
\thispagestyle{empty}
\pagestyle{empty}

\begin{abstract}

  We consider the problem of developing robots that navigate like
  pedestrians on sidewalks through city centers for performing
  various tasks including delivery and surveillance. One particular
  challenge for such robots is crossing streets without pedestrian
  traffic lights. To solve this task the robot has to decide based
  on its sensory input if the road is clear.  In this work, we propose
  a novel multi-modal learning approach for the problem of autonomous
  street crossing.  Our approach solely relies on laser and radar data
  and learns a classifier based on Random Forests to predict when
  it is safe to cross the road. We present extensive
  experimental evaluations using real-world data collected from
  multiple street crossing situations which demonstrate that our
  approach yields a safe and accurate street crossing behavior and
  generalizes well over different types of situations. A comparison to
  alternative methods demonstrates the advantages of our approach.

\end{abstract}

\section{Introduction}

The last two decades have seen tremendous advances in the fields of
mobile robotics and autonomous vehicles.
Research initiatives have been active in attempting to solve the
challenges of urban navigation such as traffic merging, navigating in
narrow lanes, and handling intersections. Initiatives such as RoboCup
and DARPA are aimed for the development of autonomous agents for
complex tasks. Over time both initiatives expanded taking significant
strides towards solving the next milestone including but not limited
to logistic robotics,
rescue robotics, and autonomous urban driving. 

We consider pedestrian robots that are designed to autonomously navigate
on sidewalks among pedestrians in urban environments and provide
assistance to users for tasks such as parcel delivery, guidance or surveillance.
Similar to self-driving vehicles, they need to navigate within the
environment and interact with surrounding vehicles in a safe and
orderly manner.  One of the key requirements for such robots is
to properly perceive their environment. In order to ensure safe operation,
the agents should be able to identify possibly dangerous situations
and seek a plan that avoids them. 

For both pedestrian robots and autonomous vehicles, street intersections
pose a threat not only for them but also for surrounding traffic if
not handled correctly.  Approaches aiming to solve this challenging
problem depend on the type of intersection, whether a traffic light
regulated one, an unsignalized crossing or a zebra crossing.
Detecting and recognizing traffic lights in a scene is a difficult
problem due to the small size of the light, the presence of similarly
colored objects in the scene and especially in the case of self-driving
vehicles, a decision needs to be made almost instantaneously.  Even if
a traffic light is detected correctly it is always desirable to
determine if it is safe to cross, e.g., when a speeding car or
an ambulance is approaching.

For unsignalized crossing situations the problem is even harder.
Without a clear signal, such as a traffic light, the robot must make a
decision based on the behavior of surrounding vehicles. Current
approaches to solve this problem employ the use of a
vehicle-to-vehicle communication systems alongside with a reservation
based approach~\cite{rios2016survey}. However such approaches were
developed with automated vehicles in mind, and would require equipping
all vehicles with such a system to function properly.


\begin{figure}
        \centering
        \includegraphics[width=0.95\columnwidth,height=5cm,keepaspectratio]{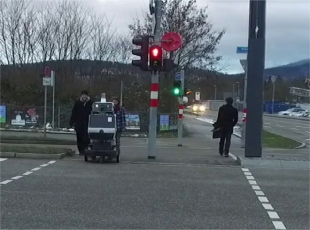}
        \caption{Autonomous navigation at a street intersection by combining
        information from different modalities. The figure shows our robotic
        platform at a street intersection, where it uses the data from
        the laser and radar sensors to decide when it is safe to cross. 
        }
        \label{fig:overview}
\end{figure}

In this paper, we present a novel approach that addresses the problem
of autonomous road crossing.  We consider pedestrian robots operating on
sidewalks and employ a multi-modal approach to solve the problem of
perceiving vehicles in traffic (see~\figref{fig:overview}).  Our robot
is equipped with two electronically scanning radars (ESR) on the sides
to cover long distances and as well as with multiple laser range
scanners to perceive the vicinity of the robot.  We compute tracks of
objects in laser range data and use radar detections in the form of
radial velocity, distance and approach angle over a fixed time
interval.  Using data labeled by humans, we train a Random Forest
classifier to predict when it is safe to cross a street.  In parallel,
we introduce a real-world dataset that we make publicly available.
The data was collected from different intersections in Freiburg,
Germany over the course of two weeks, and contains over $1,200$
annotated scenes of different crossing scenarios.  We evaluate our
approach on this dataset and show that our algorithm is able to
accurately determine safe situations. Furthermore, we compare the
prediction accuracy and generalization capabilities of the Random
Forest classifier with different classifiers and a baseline approach.
Extensive experimental evaluations demonstrate the performance gain
using the proposed method.

\section{Related Work}

Research on the problem of safe autonomous navigation across
intersections has been an active topic in the context of self-driving
vehicles. However, few approaches have addressed this problem for
pedestrian robots.  \citeauthor{bauer2009autonomous} present an
autonomous outdoor navigation pedestrian robot operating in outdoor urban
environments~\cite{bauer2009autonomous}. The robot is capable of
navigating through signalized crossings by detecting and classifying
traffic lights and signals at the
intersection. \citeauthor{baker2005automated} use a vision based
system for autonomous street crossing targeted at assistive
robots~\cite{baker2005automated}.  Using cameras mounted to the left
and right of the platform, they track oncoming vehicles to determine
whether it is safe to cross. Once the decision to cross the street has
been made, they continue to track oncoming vehicles to maintain an
updated measure of the intersection safety.  Due to the short range of
the camera, the approach can only detect nearby vehicles in a two-lane
street.

In the context of autonomous vehicles, the problem has been divided
into smaller sub-problems depending on the intersection type. For
traffic light regulated intersections, state-of-the-art approaches
vary between combining computer vision-based methods with prior scene
information to improve detection accuracy while reducing the search
space within the image~\cite{jie2013new, fairfield2011traffic,
  barnes2015exploiting, shi2016real}.

Navigation through unsignalized intersections such as roundabouts is a
more difficult problem.  In such scenarios, the behavior of the
vehicle is dependent on the action of surrounding agents. Several
approaches have targeted the area of vehicle coordination to enable
smooth interactions in intersection and merging
scenarios~\cite{rios2016survey}.  \citeauthor{de2013autonomous}
present a decentralized solution for intersection crossing where local
state constraints are used to enforce collision
avoidance~\cite{de2013autonomous}.
Similarly,~\citeauthor{lee2012development} develop a Cooperative
Vehicle Intersection Control (CVIC) system designed to find safe
driving trajectories for all vehicles approaching the
intersection~\cite{lee2012development}.

Recently, approaches have been developed that tackle the problem
independent of the type of intersection and without the use of
vehicle-to-vehicle communication systems~\cite{wei2013towards,
  dickmann2015making}.  \citeauthor{dickmann2015making} augment their
vehicle setup with additional radar sensors mounted to the left and
right, to enable information processing in roundabout and intersection
scenarios.  They process the raw radar information to build a tracker
that is able to detect and track oncoming vehicles in roundabout
crossings.

In contrast to the learning-based approach that we employ, some
techniques have alternatively modeled this problem in the context of
behavior prediction. Such approaches aim to compute the possible paths
that an agent might follow starting from its current
state~\cite{broadhurst2005monte, ferguson2008detection,
  hardy2013contingency}.  \citeauthor{meissner2014intersection} use a
multi-sensor tracking system for classification of relevant
objects~\cite{meissner2014intersection}.  This information is used to
predict the motion of relevant objects which facilitates decision
making in crossing scenarios.

The problem of safe navigation across intersections for mobile robots
is quite different than for autonomous vehicles, as mobile robots do
not need to make on the spot decision to stop or go. On the contrary,
the robot can stand at the pedestrian crossing until the intersection
is clear, whereas such behavior for an autonomous vehicle would be
dangerous for oncoming traffic. Furthermore as pedestrian robots do not
have the infrastructure to communicate with surrounding vehicles, they
must rely solely on the sensory information to make their decision.

Unlike behavioral prediction techniques, we do not attempt to forecast
the motion of surrounding agents or introduce environment specific
information such as number of lanes or road curvature. We are only
interested in making a binary decision as to whether or not it is safe
to cross the road. To the best of our knowledge, we are the first to
present an autonomous street crossing approach for a pedestrian robot
that uses automotive radars and laser scanners. The use of this
sensory setup enables us to monitor oncoming traffic and make an
informed decision based on vehicles up to $100~m$ away.

\section{Learning to Cross the Street}

\begin{figure}
\centering
\includegraphics[width=0.4\textwidth]{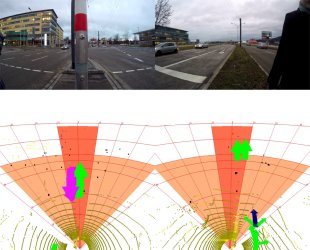}
\caption{Example input data to our approach, the top images were captured by cameras mounted to
the left and right of the platform, while bottom images display a combined visualization of
the radar and laser data. Oncoming vehicles are detected, tracked and visualized in the form of arrows.}
\label{fig:inputData}
\end{figure}

\begin{figure}[!tbp]
  \centering
  \subfloat[$t_1$]{\includegraphics[width=0.2\textwidth]{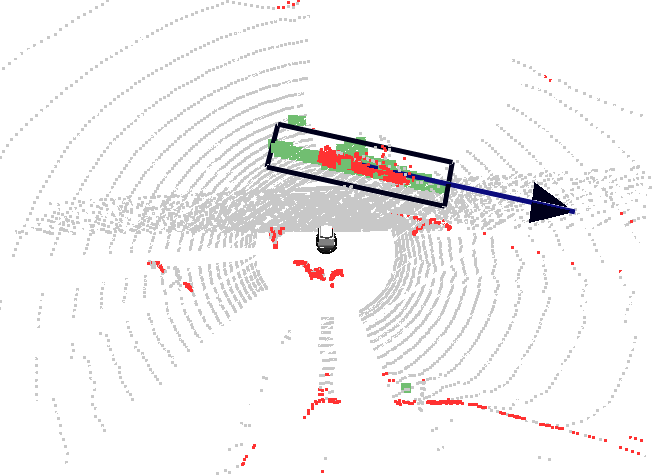}\label{fig:blob1}}
  \hspace{4pt}
  \subfloat[$t_2$]{\includegraphics[width=0.2\textwidth]{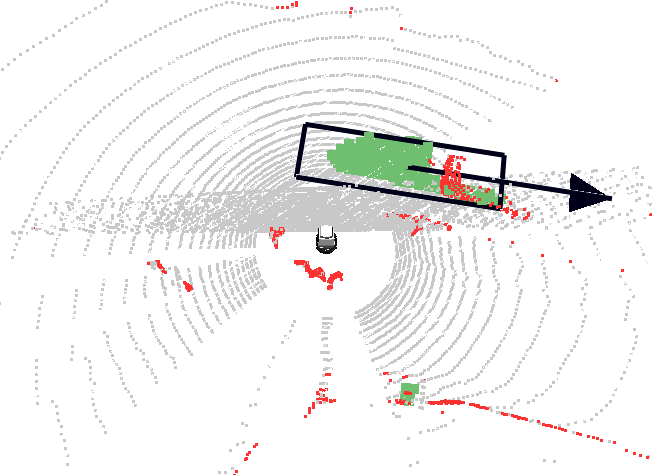}\label{fig:blob2}}\
  \caption{Visualization of the output from the dynamic obstacle
    detection approach used from two consecutive time
    intervals. Tracked objects are visualized by a surrounding
    bounding box and an arrow displaying the direction of motion.}
  \label{fig:blobTracker}
\end{figure}

We formulate the problem of safe autonomous street crrossing as a
binary classification task. The input to the classifier is the
sensor data from the most recent $\mathit{K}$-second interval, while
the output is a binary value that tells the robot as to whether it is
safe or not to cross the street. The bottom two pictures of
\figref{fig:inputData} show the typical sensory input of the radar
sensors installed on our robot. They point perpendicular to the left
and to the right of the movement direction of the robot. For
visualization purposes, the top row show images recorded with cameras
pointing in the same direction as the corresponding radar sensors. In
addition, our approach uses laser data that are preprocessed using an
object tracker developed by~\citeauthor{kummerle2015autonomous}~\cite{kummerle2015autonomous}. This
approach clusters obstacles and provides bounding box information
regarding their position, size and velocity (see
\figref{fig:blobTracker}).

Given the described hardware setup with both radar and laser sensors,
we track nearby objects $O_1, \cdots, O_m$ for a certain time interval
$\mathit{K}$. We extract features from the selected interval to create
a feature vector $\mathit{F}$, which along with a label $\mathit{L}$
is used to train a classifier. We represent each object by its ID $i$,
distance from the robot $r_i$, velocity with which it is approaching
the robot $v_i$, and the angle with which it was detected $\alpha_i$.
Accordingly, an object is represented as a triple
$\mathit{O}^T = (r, v, \alpha)$.


We create a feature vector for each time interval with a size of
$m\times n\times k$ where $n$ is the number of features for each
object (in our case $n=3$), and $k$ is the size of interval. Objects
are arranged in the vector with respect to their detection time,
followed by distance to the robot with the closest object first. Under
this representation, the final feature vector has the following
format:

\begin{eqnarray}
F = 
\begin{pmatrix}
    O_1^{t1} & O_1^{t2} & O_1^{t3} & \cdots \\[6pt]
   	O_2^{t1} & O_2^{t2} & O_2^{t3} & \cdots	\\[6pt]
   	\vdots		& \vdots		  & \vdots 		& \cdots\\[6pt]
  \end{pmatrix} \nonumber
\end{eqnarray}

We pass each feature vector as a training/testing sample
to the classifier, along with the label $\mathit{L}$ ($0$ representing
an unsafe crossing situation and $1$ a safe one).  If fewer objects
than the maximum are detected in the interval, the rest of the feature
vector is padded with zeros.

In this paper we propose a Random Forest classifier 
~\cite{breiman2001random} to learn the decision of when to cross.
We compare it to a Support Vector
Machine (SVM)~\cite{hearst1998support} approach, a k-Nearest Neighbor
(kNN)~\cite{cover1967nearest} method as well as a a baseline approach.
The latter approach iterates over all detected objects within an interval, and
independent of their temporal behavior decides if it is safe to cross
or not. For that, we use the distance of the object from
the robot and the detected velocity to compute the time to collision
assuming the velocity remains constant. If the computed time is below
a certain threshold for any of the objects throughout the time
interval, the whole interval is considered unsafe.

\section{Dataset}

\begin{figure}[!tbp]
  \centering
  \subfloat[]{\includegraphics[width=2.7cm, height=4cm, keepaspectratio]{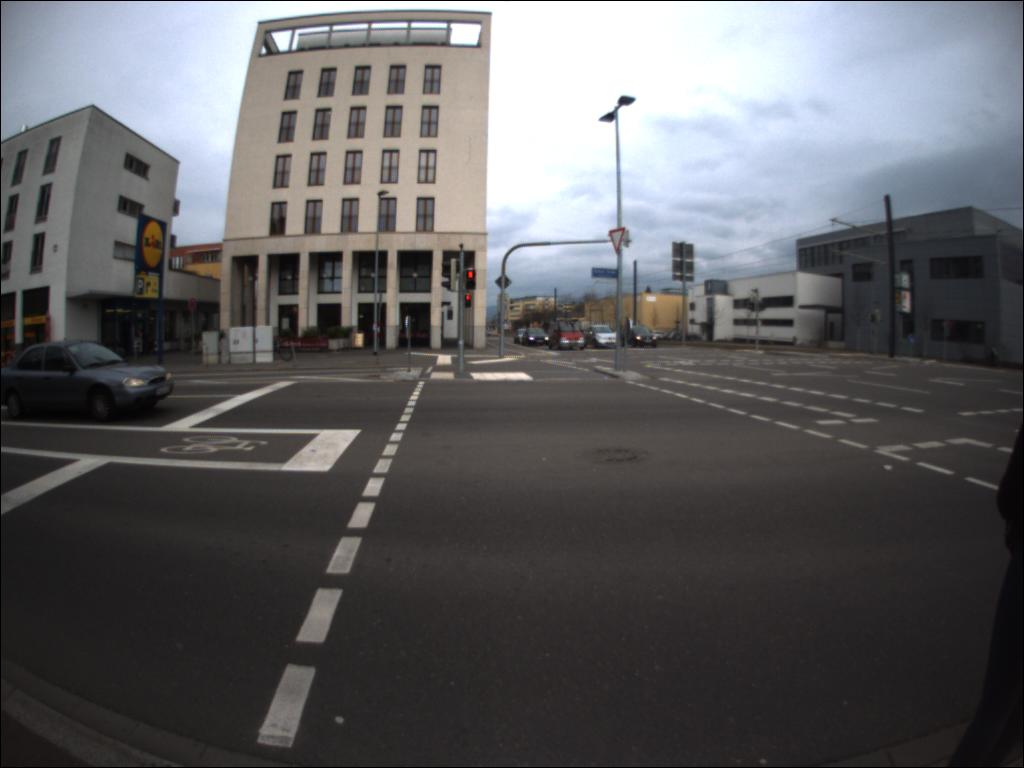}\label{fig:cross1}}
  \hspace{1pt}
  \subfloat[]{\includegraphics[width=2.7cm, height=4cm, keepaspectratio]{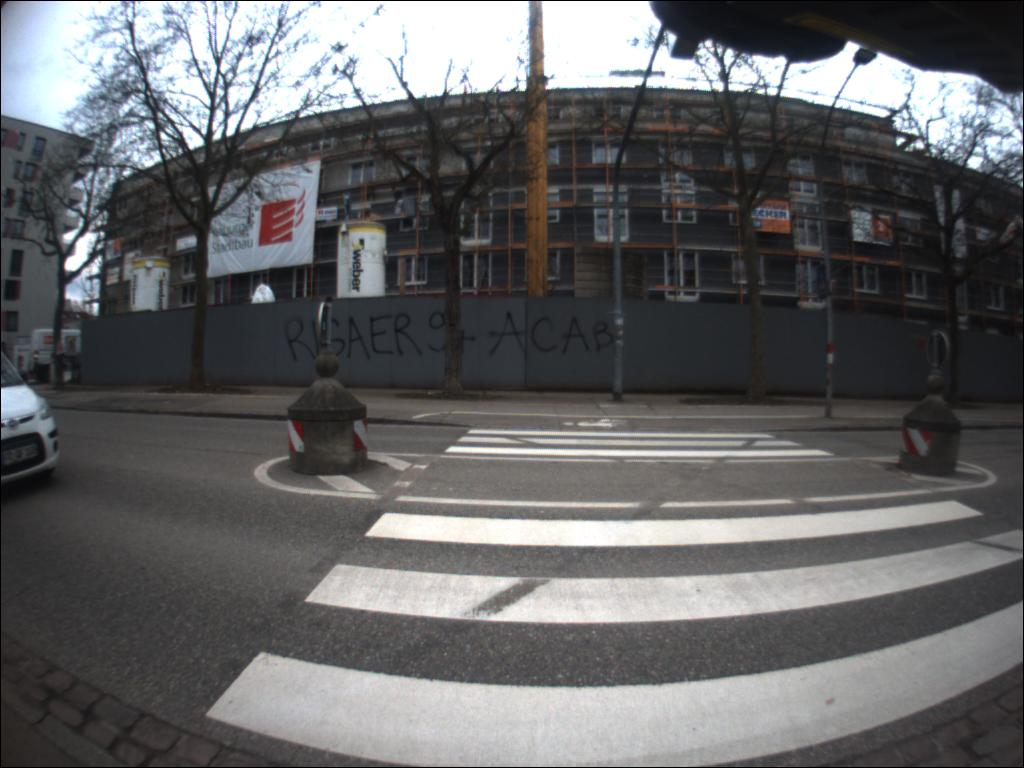}\label{fig:cross2}}
  \hspace{1pt}
  \subfloat[]{\includegraphics[width=2.7cm, height=4cm, keepaspectratio]{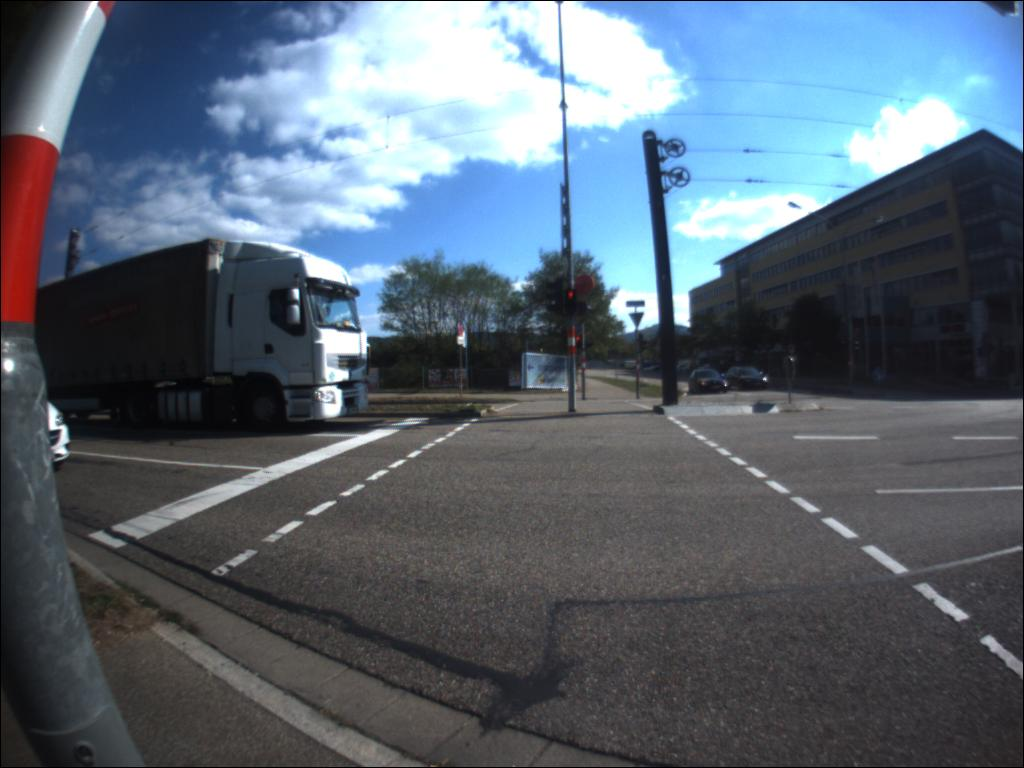}\label{fig:zebra}}
  \caption{Images from the different streets where the data was
    recorded. Both situations are shown in \protect\subref{fig:cross1}
    and \protect\subref{fig:cross2} and have traffic lights for
    pedestrians and an island in the middle. The intersection in
    \protect\subref{fig:zebra} is a zebra crossing without a middle
    island.}
  \label{fig:crossings}
\end{figure}

We collected data from three different street crossings in Freiburg,
Germany; two of which were traffic light regulated intersections and
one a zebra crossing without traffic lights. \figref{fig:crossings}
shows example images of the different intersections captured from the
perspective of the robot.  The data was gathered on different days at
different times over the course of two weeks. To collect the data, we
placed the robot on the side of the road facing the street, and
recorded live traffic data from both sides of the street. Both
traffic light regulated intersections contained an island in the
middle of the road, therefore we also recorded data standing in the
middle island facing each possible crossing direction.

The dataset used for this paper is publicly
available\footnote{\url{http://www2.informatik.uni-freiburg.de/\%7Eradwann/freiburg_street_crossing_dataset.html}}.
For each separate data file, we provide the output from the laser and
radar trackers, along with the annotation and camera images captured
from the frontal view of the platform.

Annotating the data proved to be a rather challenging problem for multiple reasons. First, the decision
to cross or not must be made using only the information from this time interval without any knowledge of
future or past intervals. Second, the time period for which
an individual observes oncoming traffic before making a decision varies from one person
to the other, rendering it difficult to assign a predetermined fixed value for it.
In addition, depending on the traffic flow people often change their decision of crossing on the spot.	
Finally, different individuals have different crossing behaviors; in the same situation at an intersection,
some might decide to cross while others choose a more conservative approach and wait for the next opportunity.
Adding more difficulty to the problem, the crossing behavior varies within the same person depending on the type of intersection and
the width of the street. These factors combined made the labeling procedure a rather tedious task, where
we attempted to eliminate as much non-determinism as possible in order to enable our classifier to learn a meaningful
classification strategy as close to human behavior as possible.
For each data sample, the decision to cross is made at the end of the interval. We used
a graphical user interface that combined the radar and laser views. Furthermore, we do not take into account
any information regarding the intersection for making the decision, e.g. number of lanes, road curvature.
Produced labels were saved with their corresponding data point to be used for the classifier training.

\section{Experiments}

\subsection{Hardware Setup}

Our robotic platform Obelix~\cite{kummerle2015autonomous} is equipped with several
sensors. For this paper, we only relied on the three laser scanners, a
Velodyne HDL-32E scanner, a tilting Hokuyo and a vertically mounted
SICK scanner. In addtion, we employed two Delphi ESRs radar sensors
which are mounted to the left and right sides of the robot. Each radar
provides both wide angle coverage at mid-range and high resolution
coverage at long-range. The radar is designed specifically for the
automotive industry, allowing the detection and tracking of adjacent
vehicles and pedestrians across the width of the equipped
platform. The long-range coverage can identify vehicles up to $174m$
with a field of view $\pm10\deg$, while the mid-range coverage has a
shorter range of only $60m$ but with a much larger field of view
$\pm45\deg$. Each radar provides tracked object information such as
time at which the object was detected, object ID, range, radial
velocity, radial acceleration and angle.

\begin{figure}
\centering
\includegraphics[width=0.4\textwidth]{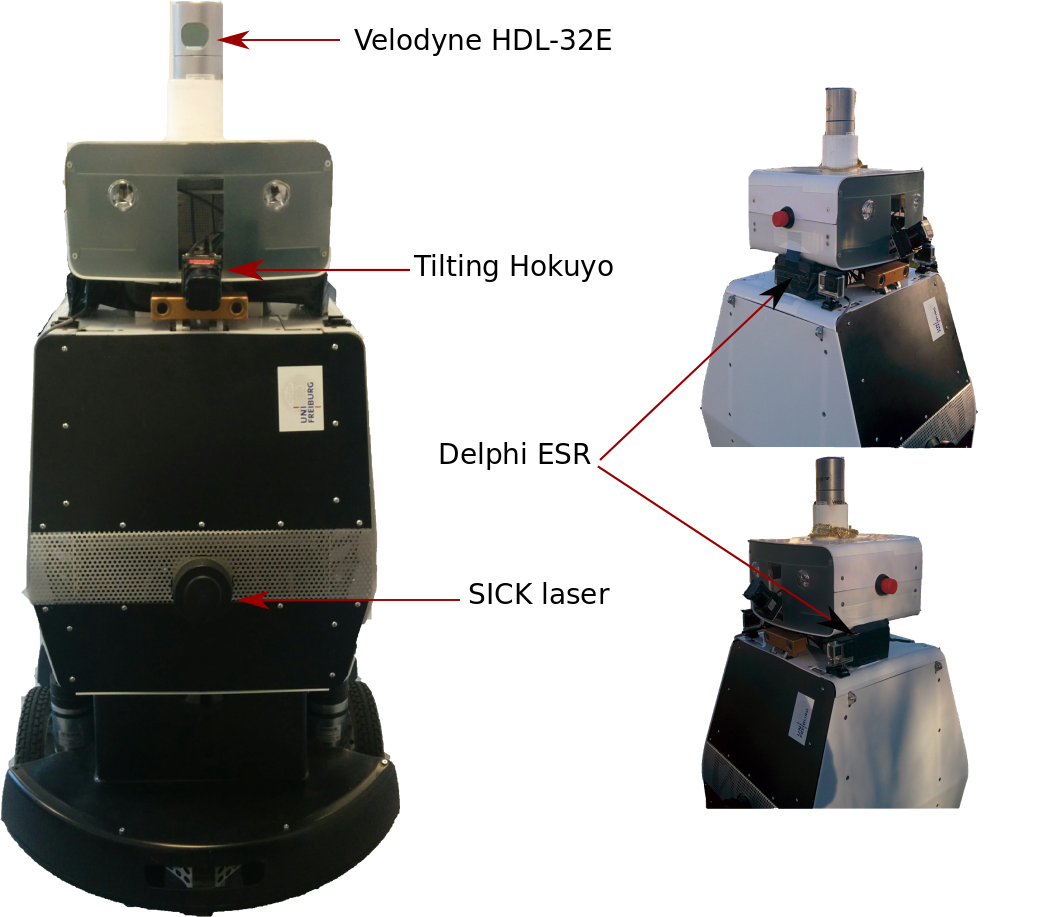}
\caption{The robot Obelix equipped with three laser scanners and two radars.}
\label{figs:obelix}
\end{figure}

\subsection{Experimental Setup}

Following the data collection procedure, we divided each file into
five second interval blocks. The resulting samples from different days
were combined forming approximately 1,270 data points.  We divided
the data into a training and a test set with a $3:2$ split ratio.  The
collected sample data shows a slightly biased class distribution, with
more non crossing examples than crossing ones with a ratio of $5:4$.

The parameters of the classifiers were selected by an exhaustive grid
search over the hyperparameter space.  We evaluated each parameter
setting on the training data by applying
leave-one-out-cross-validation.  The training data was divided into
five folds of equal size. We used four folds for training the model
and one for evaluating the current parameter configuration. In the
remainder of this paper, we report the results using the parameter
configuration producing the best average precision and recall values.
For the baseline classifier, the minimum time to collision threshold
was set to $10\mathit{sec}$. This value was selected as a
representative of the average time it took our platform to cross from
one end of the sidewalk to the other.

\subsection{Quantitative Results}

In this section, we present different quantitative measures for
comparing the proposed approach.  Evaluating the different parameter
configurations on the performance of the Random Forest classifier
shows that the learning behavior of the classifier is robust to the
selected parameters. We opted for a maximum tree depth of $100$, a
minimum sample size of $50$ and an active variable size of $100$.  We
evaluated the performance of the SVM classifier on the kernel type,
$\mathit{C}$-value and $\gamma$-value, and obtain the best performance
using a Sigmoid kernel with a $\mathit{C}$-value of $2.0$ and a
$\gamma$-value of $0.1$. Cross validation showed that changing the
values of either parameter did not lead to a significant performance
improvement.  For training the kNN classifier, we used a
$\mathit{k}$-value of $8$, which proved to provide the best compromise
between precision and recall.

\begin{figure}[!tbp]
  \centering
  \subfloat[Random Forest]{\includegraphics[width=0.25\textwidth]{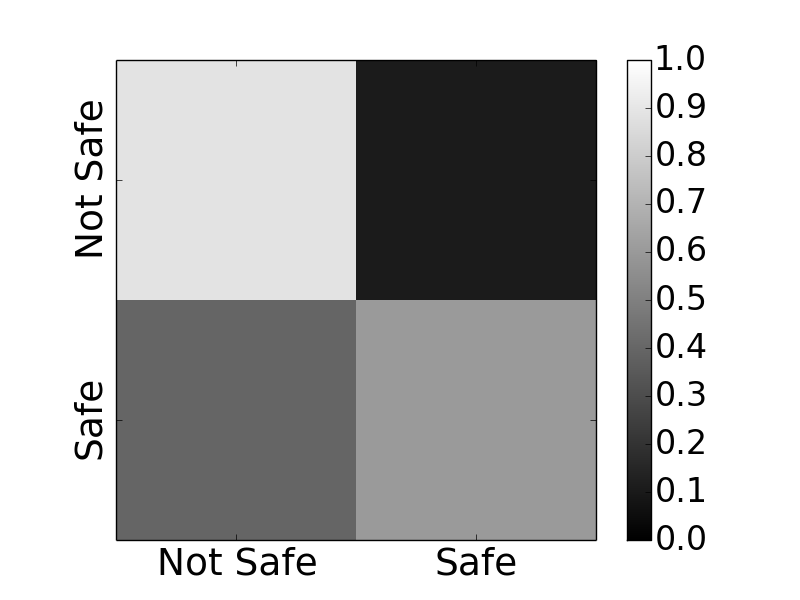}\label{fig:rt}}
  \subfloat[SVM]{\includegraphics[width=0.25\textwidth]{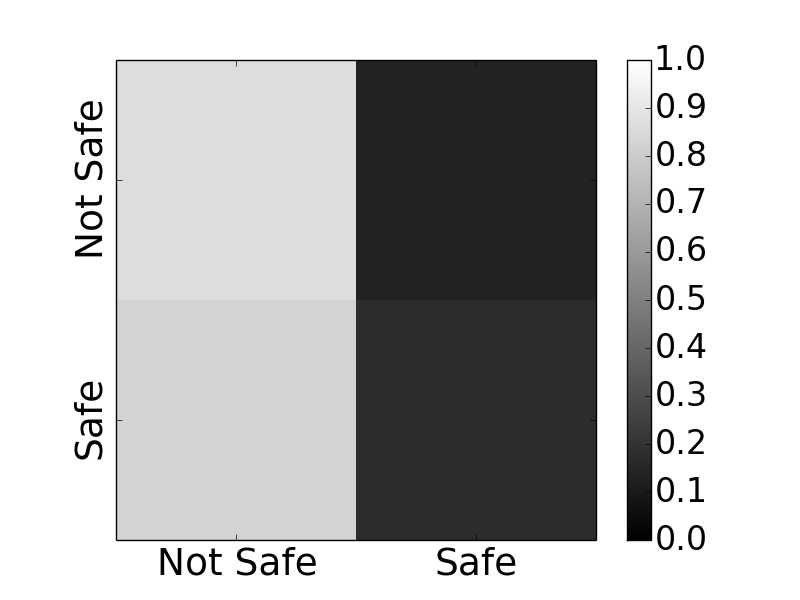}\label{fig:svm}}\\
  \subfloat[kNN]{\includegraphics[width=0.25\textwidth]{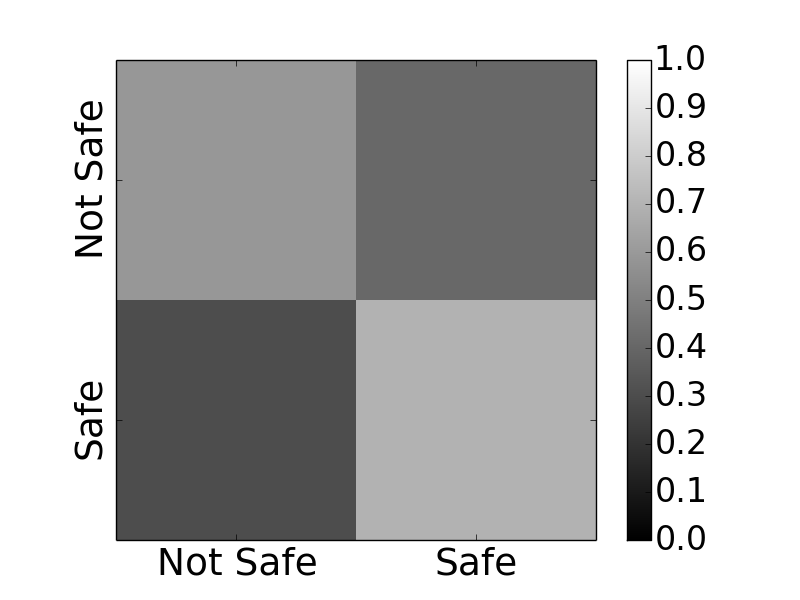}\label{fig:knn}}
  \subfloat[Baseline]{\includegraphics[width=0.25\textwidth]{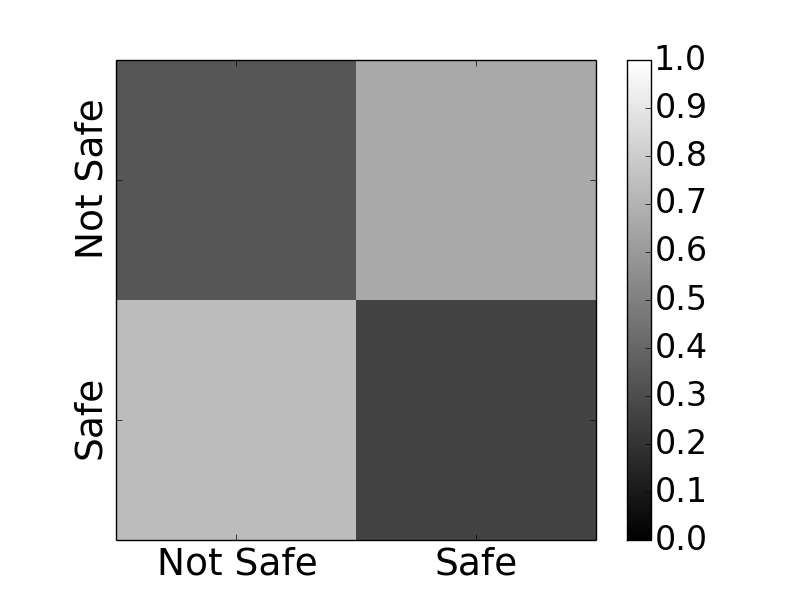}\label{fig:baseline}}
  \caption{Confusion matrix for the various trained classifiers on the test dataset. The Random
  Forest classifier has the highest accuracy followed by the kNN classifier.}
  \label{fig:confMats}
\end{figure}

\begin{figure}
\centering
\includegraphics[width=0.95\columnwidth]{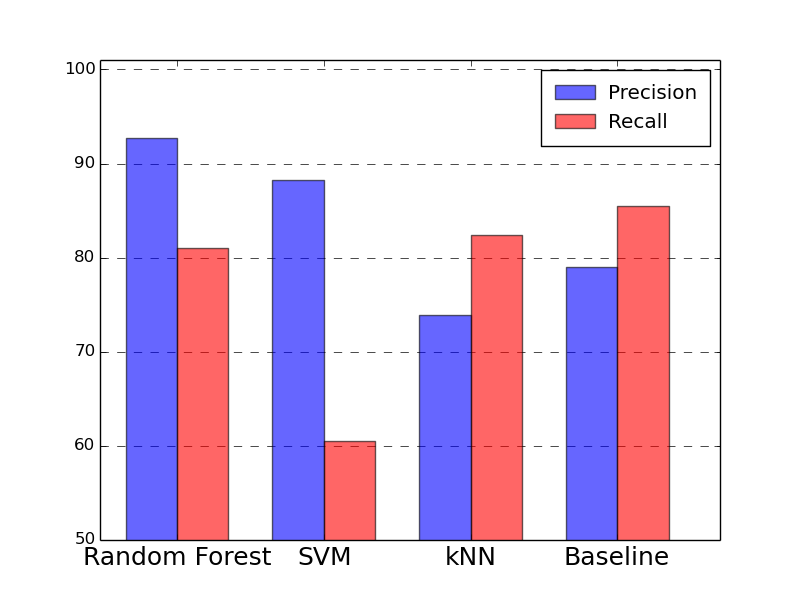}
\caption{Bar plot showing the precision and recall of the trained
  classifiers on the test data. Note that the y-axis of the plot
  starts from $50$ to better highlight the differences between the 
  classifiers. Our Random Forest classifier has the
  highest combined precision and recall values. The SVM classifier
  achieves the second highest precision but the lowest recall. On the
  other hand, the baseline classifier has the highest recall and low
  precision.}
\label{fig:performance}
\end{figure}

\figref{fig:performance} plots the precision and recall performance
for the evaluated classifiers. In our problem setting, low precision
is more dangerous than low recall, as low precision increases the risk
of getting run over by an oncoming vehicle.  On the other hand, low
recall demonstrates a more conservative crossing approach where the
robot would rather wait for the obstacle to pass than
cross. Nonetheless, we do not encourage a very conservative approach
as it could lead to the robot being caught in a deadlock situation
unable to cross. Given our problem definition, the Random Forest classifier
shows the best performance as it is able to balance between accurately
determining when to cross the street and minimizing the waiting time
for crossing.


The confusion matrix for the different classifiers is shown
in~\figref{fig:confMats}. Our Random Forest classifier shows the best
accuracy with the lowest false positive rate in comparison to all
other classifiers. The confusion matrix of the SVM classifier shows
that it favors labeling examples as not safe to cross over safe, which
indicates that the learned classifier is more likely to wait for
longer periods of time. The kNN classifier shows slightly better
performance compared to the SVM, but with a higher number of false
positives and false negatives in comparison to the Random Forest. The
baseline approach shows the worst accuracy, consistently confusing
both classes.

\subsection{Generalization Capabilities}

\begin{figure}
\centering
\includegraphics[width=0.95\columnwidth]{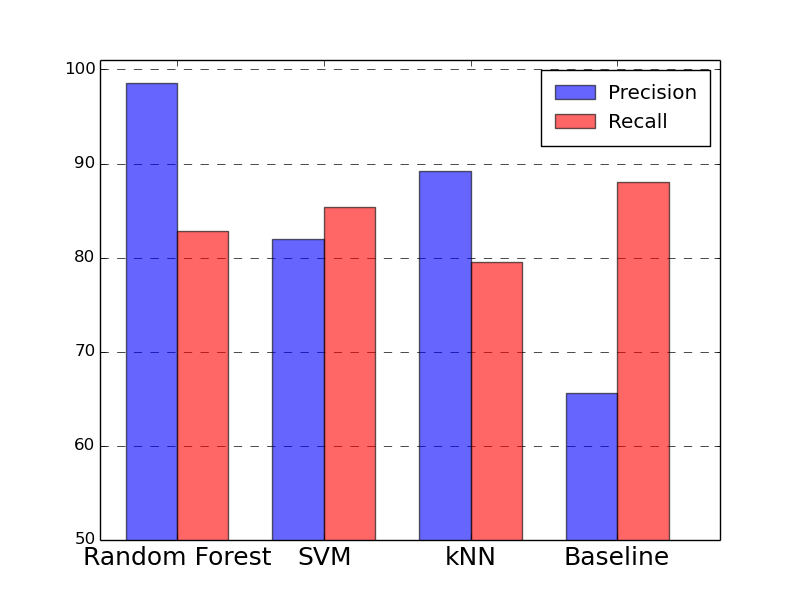}
\caption{Bar plot showing the precision and recall values of the
  evaluated classifiers, which were trained on data from two street
  crossing places and tested on sample data from a third place. 
  Note that the y-axis of the plot
  starts from $50$ to better highlight the differences between the 
  classifiers. Our
  Random Forest classifier shows the best generalization capabilities
  with the highest combined precision and recall values, $98.6\%$ and
  $82.8\%$ respectively. The lowest precision is achieved by the
  baseline classifier with a value of $65.6\%$.}
\label{fig:genPR}
\end{figure}

The goal of this experiment is to measure the robustness of the
presented approach with regard to the corresponding scenario. As
previously mentioned, we collected our data from three different
intersections. We used data from traffic light regulated
intersections to create the training set, while we test the performance
of the learned classifiers on data from the zebra crossing. After splitting
the training data was observed to have the same class distribution
as the original experimental setup, however the 
test data has a class distribution with more negative examples.
%
%

\figref{fig:genPR} displays a bar plot of the precision and recall
values for the trained classifiers. Our Random Forest classifier shows
the best generalization capabilities with a precision value of
$98.6\%$ and recall of $82.8\%$. The baseline classifier on the other
hand, shows the worst performance with a precision value slightly
better than random guessing.  The SVM classifier has higher recall
values in comparison with the initial setup, which we attribute to the
unbalanced class distribution of the test set. On the contrary, the
kNN classifier is able to generalize better with a $10\%$ improvement
in precision and a $3\%$ improvement in recall relative to the initial
setup.

\subsection{Qualitative Results}

\begin{figure*}[!tbp]
  \centering
  \subfloat[$\mathit{time}=0~sec$]{\includegraphics[width=0.3\textwidth]{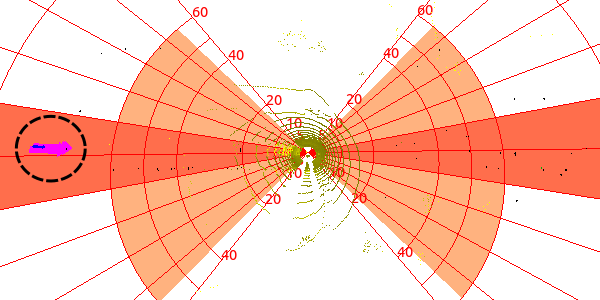}\label{fig:seq1}}
  \hspace{4pt}
  \subfloat[$\mathit{time}=2.5~sec$]{\includegraphics[width=0.3\textwidth]{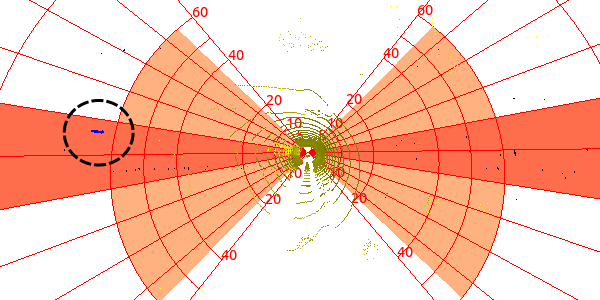}\label{fig:seq2}}
  \vspace{4pt}
  \subfloat[$\mathit{time}=5~sec$]{\includegraphics[width=0.3\textwidth]{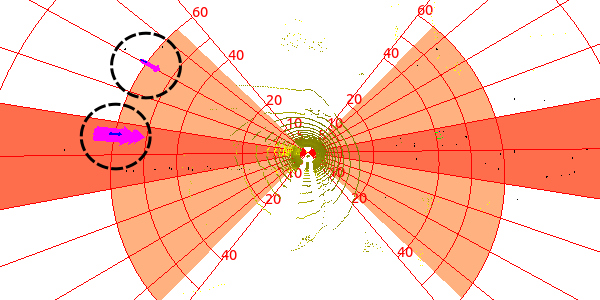}\label{fig:seq3}}
  \caption{Example of a misclassified sample from the test data, with
    a safe to cross label versus an unsafe to cross prediction. 
    The dashed circle shown in the images highlights the positions of oncoming
    vehicles.
    At the
    first half of the interval a car appears to slow down, but at the
    last second it speeds up again. In such cases, the label selected
    during annotation varies greatly from one individual to the
    other.}
  \label{fig:falseNeg1}
\end{figure*}

\begin{figure*}[!tbp]
  \centering
  \subfloat[$\mathit{time}=0~sec$]{\includegraphics[width=0.3\textwidth]{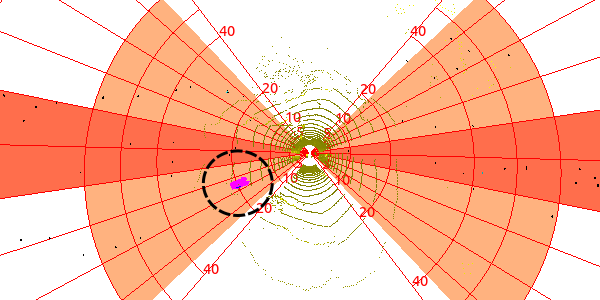}\label{fig:seq21}}
  \hspace{4pt}
  \subfloat[$\mathit{time}=2.5~sec$]{\includegraphics[width=0.3\textwidth]{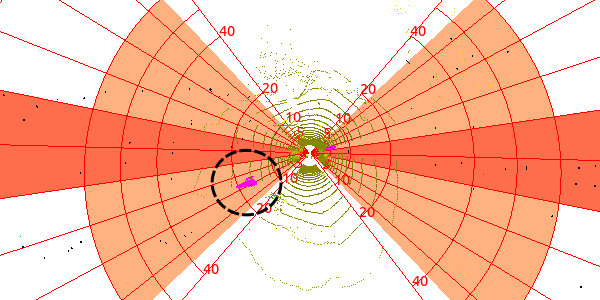}\label{fig:seq22}}
  \vspace{4pt}
  \subfloat[$\mathit{time}=5~sec$]{\includegraphics[width=0.3\textwidth]{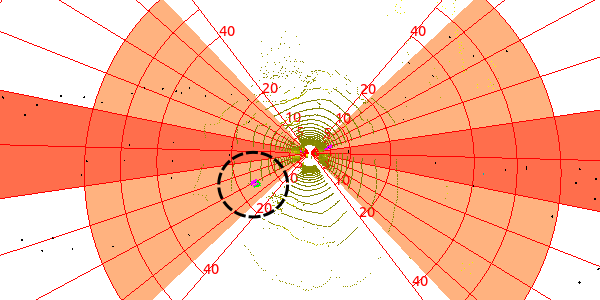}\label{fig:seq23}}
  \caption{Visualization of a false negative classification on the
    test data.
    Throughout the entire interval the car appears to be
    slowing down. However, the classifier opts for a more conservative
    approach than the annotation.}
  \label{fig:falseNeg2}
\end{figure*}

\begin{figure*}[!tbp]
  \centering
  \subfloat[$\mathit{time}=0~sec$]{\includegraphics[width=0.3\textwidth]{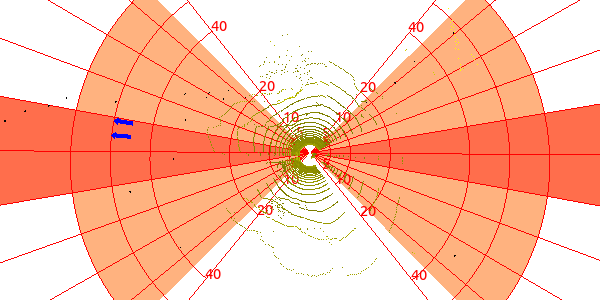}\label{fig:seq31}}
  \hspace{4pt}
  \subfloat[$\mathit{time}=2.5~sec$]{\includegraphics[width=0.3\textwidth]{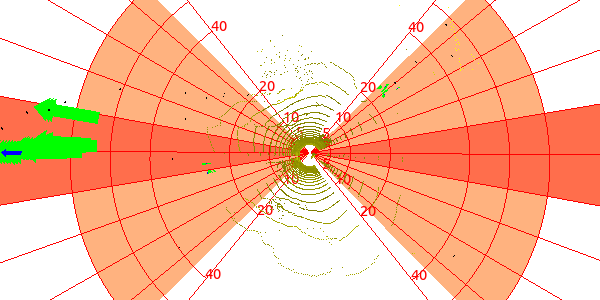}\label{fig:seq32}}
  \vspace{4pt}
  \subfloat[$\mathit{time}=5~sec$]{\includegraphics[width=0.3\textwidth]{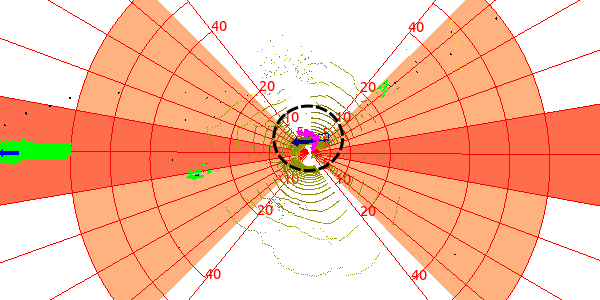}\label{fig:seq33}}
  \caption{Example of an incorrect classification on a data point. 
  The
    classifier predicted the interval to be safe to cross, whereas at
    the last second in the window a car is passing in front of the
    robot.  Both trackers were unable to detect the car until the last
    second as it was outside of their sensing range due to the
    curvature of the road.}
  \label{fig:falsePos}
\end{figure*}

In this section, we present qualitative results on the presented
dataset trained using the Random Forest classifier.
\figref{fig:falseNeg1} and \figref{fig:falseNeg2} show examples of
false negative predictions by the classifier.
In~\figref{fig:falseNeg1}, the tracked vehicle slows down during the
first half of the interval, then continues to speed up again in the
remainder of it. In this example the decision of crossing varies
within individuals, making it difficult to define what a ground-truth
label should be.  \figref{fig:falseNeg2} shows an example scenario in
which a car appears to be approaching during the length of the
interval with decreasing velocity. Ideally, we expect our classifier to
learn in such cases that it is safe to cross, mimicking the behavior
of humans at zebra crossings. However, the trained classifier is
unable to learn such a behavior due to the insufficient number of
examples showing similar situations in the training data. On the other
hand,~\figref{fig:falsePos} demonstrates a situation where the learned
classifiers labels an unsafe situation as safe. In this example, a car
coming from a side street outside of the sensor range of the robot is
treated by the classifier as a false tracker detection leading to
incorrect safe classification. We believe this occurs because the car
appears at the very end of the interval and only for a few seconds,
which closely resembles the characteristics of a ghost detection.

\section{Conclusions}

In this paper we presented a novel approach based on Random Forests
for learning to predict when it is safe to cross a street.  We employ
a pedestrian robot that uses multiple sensor modalities. Our approach
takes into account information from laser and radar sensors to detect
moving objects. Given such data over a short time interval, it decides
whether it is safe to cross the street or not. We trained and
evaluated a Random Forest classifier based on these modalities using
real-world data from different places. The corresponding dataset has
been made publicly available. The experimental results show that our
classifier is robust to the type of intersection, generalizes well to
different situations and outperforms different alternative approaches.

\addtolength{\textheight}{-5cm}   

\bibliographystyle{abbrvnat}
\footnotesize
\bibliography{bib}

\end{document}